\documentclass[letterpaper]{article} %
\usepackage{aaai23}  %
\usepackage{times}  %
\usepackage{helvet}  %
\usepackage{courier}  %
\usepackage[hyphens]{url}  %
\usepackage{graphicx} %
\usepackage{natbib}  %
\usepackage{caption} %
\usepackage{algorithm}
\usepackage{algorithmic}

\usepackage{color}

\usepackage{amssymb}
\usepackage{bm}
\usepackage{pifont}
\usepackage{amsmath}
\usepackage{mathtools}
\usepackage{etoolbox,lipsum}

\def\eg{\textit{e.g.}}
\def\ie{\textit{i.e.}}
\def\etc{\textit{etc}}
\newcommand{\redbf}[1]{{\textbf{\color{red}{#1}}}} %
\newcommand{\blueud}[1]{{\underline{\color{blue}{#1}}}} %

\usepackage{newfloat}
\usepackage{listings}
\DeclareCaptionStyle{ruled}{labelfont=normalfont,labelsep=colon,strut=off} %
\floatstyle{ruled}
\newfloat{listing}{tb}{lst}{}
\floatname{listing}{Listing}
\title{Mitigating Artifacts in Real-World Video Super-Resolution Models}
\author{
	Liangbin Xie\thanks{Liangbin Xie is an intern in ARC Lab, Tencent PCG.}$^{1,2,3}$,\hspace{2pt} 
	Xintao Wang$^{3}$,\hspace{2pt}
	Shuwei Shi$^{1,4}$,\hspace{2pt}
	Jinjin Gu$^{5,6}$,\hspace{2pt}
	Chao Dong\thanks{Corresponding author.}$^{1,6}$,\hspace{2pt} 
	Ying Shan$^{3}$
}
\affiliations {
    \textsuperscript{\rm 1} The Guangdong Provincial Key Laboratory of Computer Vision and Virtual Reality Technology, \\ Shenzhen Institute of Advanced Technology, Chinese Academy of Sciences \hspace{9pt}
    \textsuperscript{\rm 2} University of Macau \hspace{9pt} \\
    \textsuperscript{\rm 3} ARC Lab, Tencent PCG \hspace{9pt}
    \textsuperscript{\rm 4} Shenzhen International Graduate School, Tsinghua University \\
    \textsuperscript{\rm 5} The University of Sydney \hspace{9pt} 
    \textsuperscript{\rm 6} Shanghai Artificial Intelligence Laboratory \\
    \{lb.xie, chao.dong\}@siat.ac.cn \hspace{9pt}  ssw20@mails.tsinghua.edu.cn \hspace{9pt} \\
    jinjin.gu@sydney.edu.au \hspace{9pt}
    \{xintaowang, yingsshan\}@tencent.com
}

\begin{document}

\maketitle

\begin{abstract}
	The recurrent structure is a prevalent framework for the task of video super-resolution, which models the temporal dependency between frames via hidden states.
	When applied to real-world scenarios with unknown and complex degradations, hidden states tend to contain unpleasant artifacts and propagate them to restored frames.
	In this circumstance, our analyses show that such artifacts can be largely alleviated when the hidden state is replaced with a cleaner counterpart.
	Based on the observations, we propose a Hidden State Attention (HSA) module to mitigate artifacts in real-world video super-resolution.
	Specifically, we first adopt various cheap filters to produce a hidden state pool. For example, Gaussian blur filters are for smoothing artifacts while sharpening filters are for enhancing details.
	To aggregate a new hidden state that contains fewer artifacts from the hidden state pool, we devise a Selective Cross Attention (SCA) module, in which the attention between input features and each hidden state is calculated.
	Equipped with HSA, our proposed method, namely FastRealVSR, is able to achieve 2${\times}$ speedup while obtaining better performance than Real-BasicVSR. Codes will be available at \textcolor{cyan}{https://github.com/TencentARC/FastRealVSR}.
\end{abstract}

\section{Introduction}
\label{sec:intro}

Video Super-Resolution (VSR) aims to recover high-resolution (HR) frame sequences from their low-resolution (LR) counterparts, where the utilization of complementary information across adjacent frames is the key factor to improve performance.
VSR can be roughly classified into two categories -- sliding-window-based methods~\cite{caballero2017real,huang2017video,wang2019edvr,tian2020tdan,li2020mucan,isobe2020video} and recurrent methods~\cite{chan2021basicvsr,chan2022basicvsr++,huang2015bidirectional,sajjadi2018frame,haris2019recurrent,xiang2020zooming,isobe2020video,isobe2020revisiting,lin2021fdan}.
Recurrent methods are the prevalent approaches since they can model long-term dependencies with hidden states, and have won several championships in video restoration competitions~\cite{yang2021ntire,yang2022ntire}.

\begin{figure}[t]
	\centering
	\includegraphics[width=1.0\columnwidth]{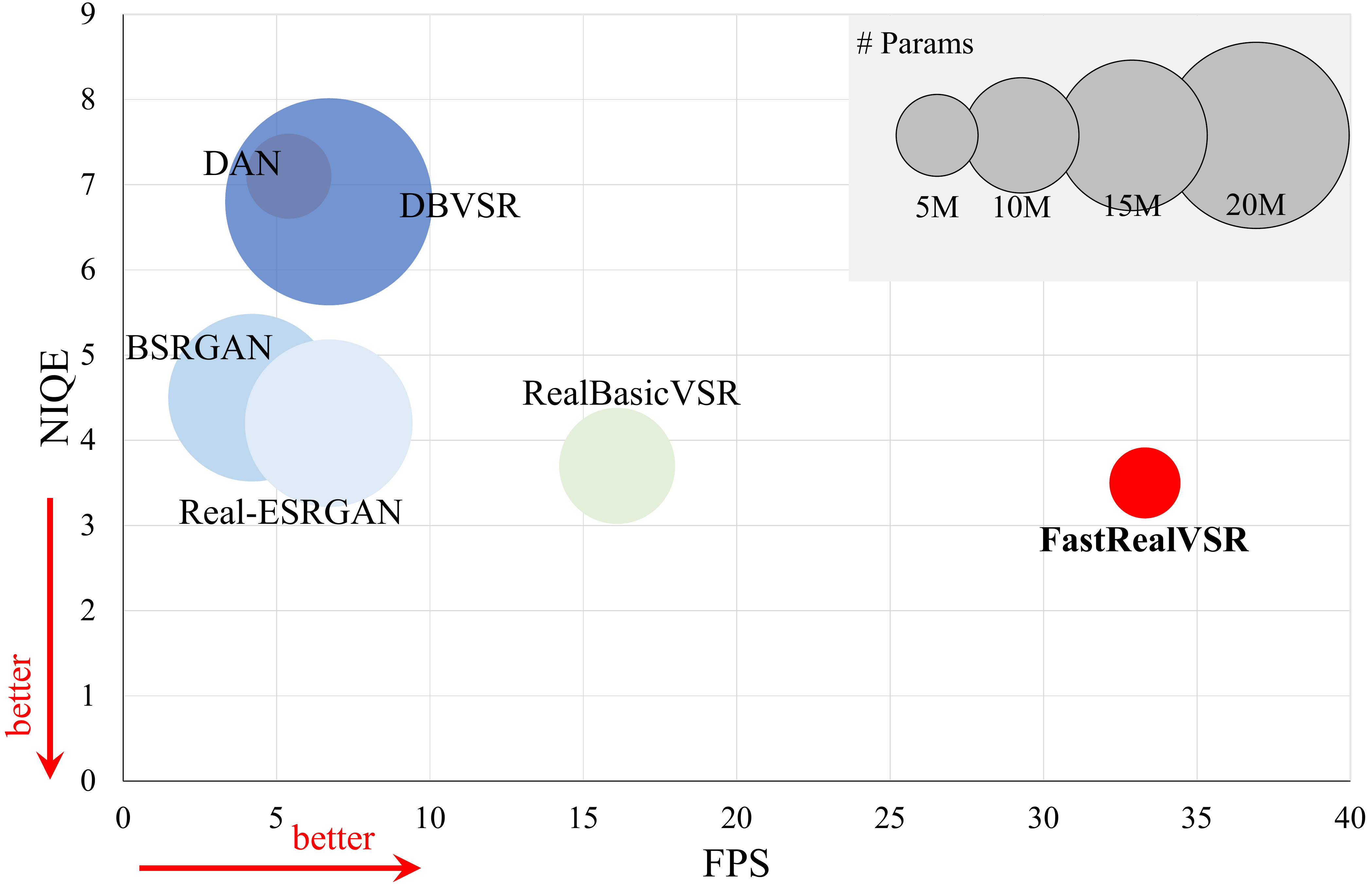}
	\caption{Our proposed FastRealVSR can ahieve the best trade-off between NIQE, speed and parameters in the real-world VideoLQ dataset.}
	\label{fig:speed_cmp}
\end{figure}

When applied to real-world low-quality videos, existing recurrent methods (\eg, BasicVSR~\cite{chan2021basicvsr}) tend to produce unpleasant artifacts, which will be amplified through propagation~\cite{chan2022investigating}. This problem becomes more severe when the degradations are more complicated with unknown blur, noise and compression.

Recent work Real-BasicVSR~\cite{chan2022investigating} makes the first attempt to alleviate those artifacts. It adopts a separate image pre-cleaning module before the video recurrent restoration model to explicitly remove degradations for each video frame.
However, while mitigating unpleasant artifacts, such a strategy can also lead to detail loss, due to the explicit denoising operation in the image level.
Besides, the separate module brings extra computation cost, which slows down the inference speed and is not friendly to practical usages.

In this work, we carefully investigate the cause of the unpleasant artifacts in VSR. We draw two observations.
1) Once the output of the previous frame produces artifacts, the hidden state will also contain such artifacts and further propagate to consequent frames. When we discard the hidden state and only use the current input frame, those artifacts disappear, but at the cost of detail loss.
2) When we replace the ``problematic'' hidden state with a cleaner one (\eg, the corresponding one from the MSE model\footnote{Training with MSE loss tends to produce over-smooth results and thus the hidden state is also cleaner and more smooth.} or a Gaussian-blurred one), the artifacts are largely mitigated and do not further propagate to consequent frames.

Such observations motivate us to directly manipulate the hidden state instead of introducing extra cleaning modules.
We propose a Hidden State Attention (HSA) module to mitigate artifacts in real-world video super-resolution.
Specifically,
we first generate a hidden state pool, where each new hidden state is produced by filtering the original hidden state with different filters, including Gaussian blur filters, box blur filters, sharpening filters, \etc.
Those operations will either blur hidden state to alleviate artifacts or sharpen edges to enhance details.
We choose those classical filters as they only bring a little computation cost.
After that, selective cross attention will calculate the similarity between the features of the current frame and each hidden state in the pool.
As ``problematic'' hidden state usually has a large deviation from input features, such a design can effectively mitigate artifacts.
A cleaned hidden state is then aggregated from the hidden state pool based on the calculated attention.
As a result, this new hidden state not only contains fewer artifacts, but also maintains the details.

\begin{figure}[t]
	\centering
	\includegraphics[width=.95\columnwidth]{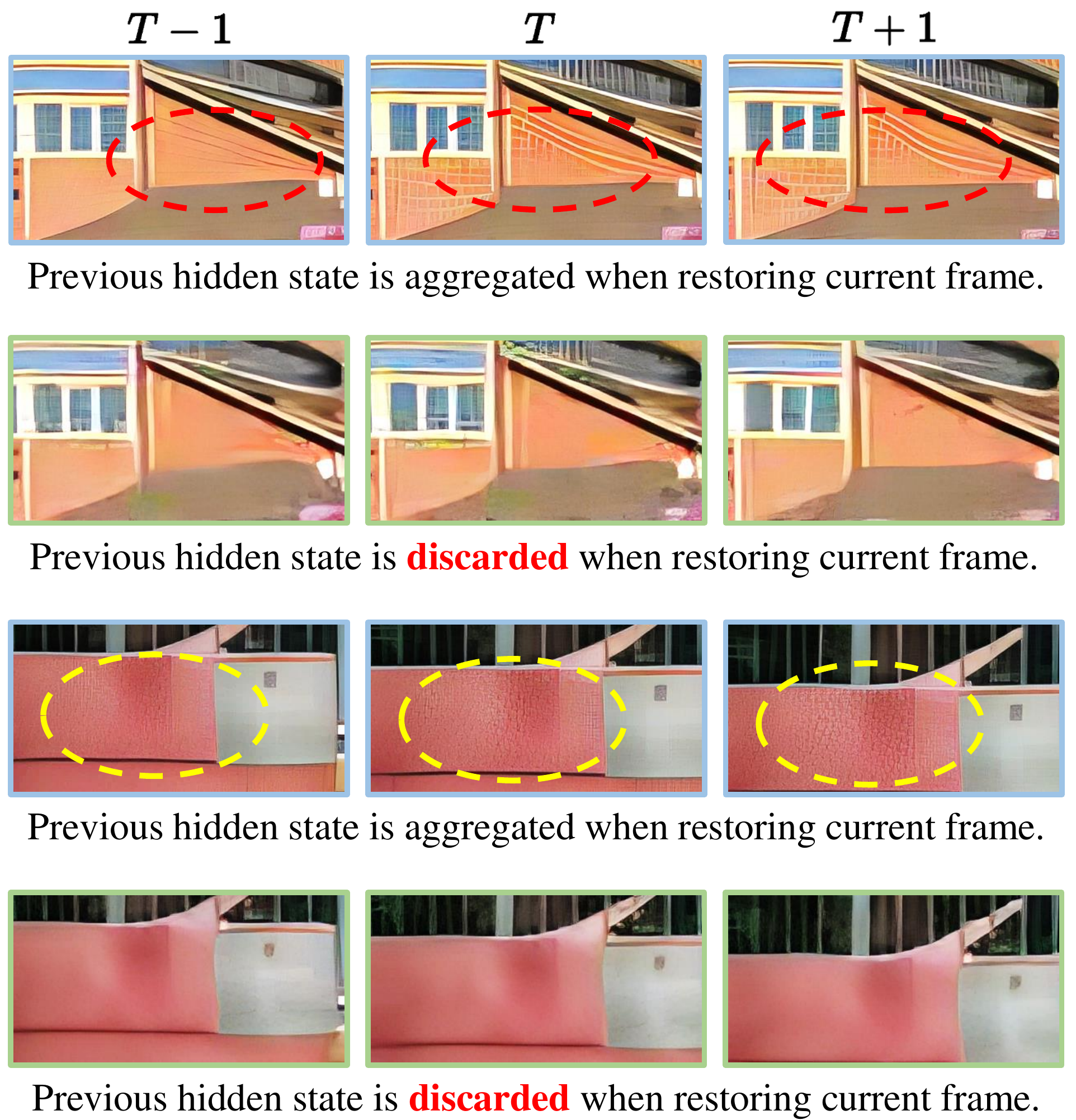}
	\caption{Negative impact of hidden state in real-world scenario. The hidden state could contain artifacts that appears in the previous frames, and further propagate to consequent frames. The regions marked with the ellipse dotted box are the artifacts. (\textbf{Zoom-in for best view})}
	\label{fig:w_wo_hidden_state}
\end{figure}

To make our algorithm more practical in the real world, we also take inference speed into consideration. We adopt a compact unidirectional recurrent framework as the backbone of our method. Together with the efficient Hidden State Attention (HSA) module, our proposed method, namely FastRealVSR, achieves $2\times$ speed-up over Real-BasicVSR, while keeping comparable performance.

Our contributions can be summarized as follows.
\textbf{1)} We analyze the hidden state in recurrent VSR methods and show that the hidden state is the main reason for unpleasant artifacts.
\textbf{2)} An effective hidden state attention with cheap hidden states and selective cross attention is proposed to mitigate artifacts in real-world VSR models.
\textbf{3)} FastRealVSR is capable of achieving a better speed-performance trade-off, surpassing previous methods.

\section{Related Work}
\label{sec:related_work}

\begin{figure}[t]
	\centering
	\includegraphics[width=1\columnwidth]{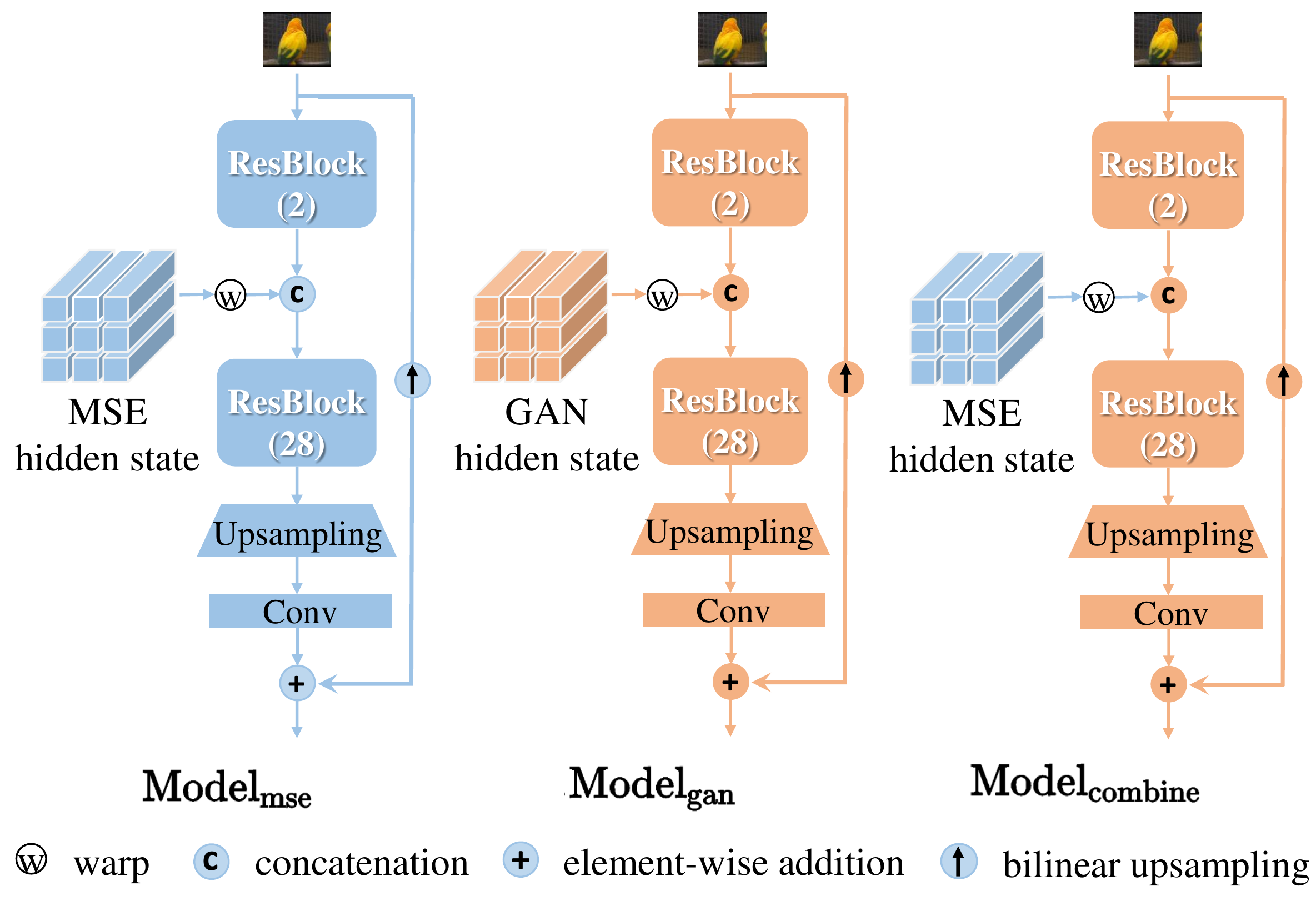}
	\caption{The diagram of general unidirectional recurrent network. $\mathrm{Model_{combine}}$ is derived by replacing the GAN hidden state in $\mathrm{Model_{gan}}$ with the MSE hidden state in $\mathrm{Model_{mse}}$. We use a single color to represent a complete pre-trained model. (\textbf{Zoom-in for best view})}
	\label{fig:inference}
\end{figure}

\textbf{Video Super Resolution.} Most existing VSR methods can be divided into two categories: sliding-window based methods~\cite{caballero2017real,huang2017video,wang2019edvr,tian2020tdan,li2020mucan,isobe2020video} and recurrent methods~\cite{chan2021basicvsr,chan2022basicvsr++,huang2015bidirectional,sajjadi2018frame,haris2019recurrent,xiang2020zooming,isobe2020video,isobe2020revisiting,lin2021fdan}. Sliding-window methods often take several adjacent frames as input and predict the center frame. In contrast to sliding-window based methods, recurrent methods adopt recurrent neural network~\cite{mikolov2010recurrent} to process the input frame and generate the output progressively. Some recent methods~\cite{chan2021basicvsr,chan2022basicvsr++} have achieved excellent performance on synthetic degradations (\eg, bicubic).
Recently, several transformer-based~\cite{cao2021video,liang2022vrt} methods are proposed. However, when these methods~\cite{chan2021basicvsr} are applied to real-world scenarios, their performance deteriorates severely and obvious artifacts will appear since the degradations in real-world scene are unknown and complex.

\noindent \textbf{Real-World Video Super-Resolution.} Recent work RealBasicVSR~\cite{chan2022investigating} is proposed to investigate tradeoffs in real-world VSR.  Similar to Real-ESRGAN~\cite{wang2021real}, a second-order degradation model is adopted to simulate the degradations in the real-world. To alleviate the artifacts that are amplified in recurrent
methods (\eg BasicVSR), it uses a pre-cleaning module and a cleaning loss to ``clean'' the input sequence, which is then fed into a general VSR network. In this paper, to mitigate the artifacts, we do not introduce any cleaning modules. Instead, we manipulate the hidden state directly to mitigate the artifacts. Compared to RealBasicVSR, our proposed FastRealVSR achieves a better speed-performance trade-off.

\noindent \textbf{Hidden State Manipulation.} Hidden state is the key component in recurrent methods that adopted in video restoration task~\cite{nah2019recurrent,zhong2020efficient}. To handle the misalignment between the hidden state and the referred current feature, some works~\cite{nah2019recurrent,zhou2019spatio} in video deblurring area are proposed to adapt the hidden states to the current frame by referring to the current feature. Different from them, we aggregate a new hidden state from a hidden state pool and aim to alleviate the mitigate contained in ``problematic'' hidden state.

\section{Methodology}
\label{sec:method}

\subsection{Analysis of Hidden State}

\begin{figure}[t]
	\centering
	\includegraphics[width=1\columnwidth]{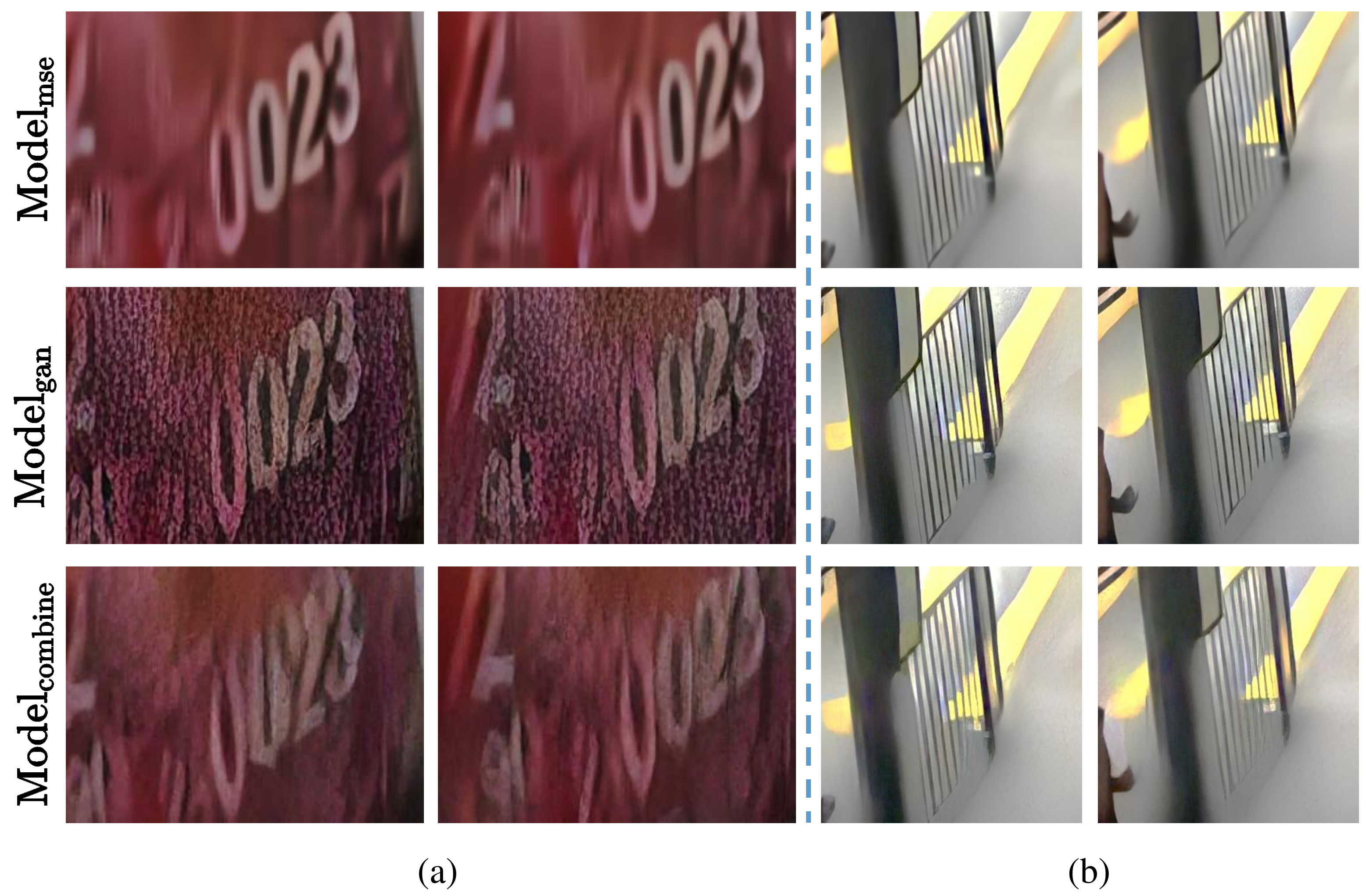}
	\caption{Qualitative comparison between $\mathrm{Model_{mse}}$, $\mathrm{Model_{gan}}$ and $\mathrm{Model_{combine}}$. As shown on the left side of the figure, after replacing GAN hidden state with MSE hidden state, the artifacts in the wall are largely mitigated. As shown on the right side of the figure, the output of $\mathrm{Model_{combine}}$ also loses many necessary details in the railing. (\textbf{Zoom-in for best view})}
	\label{fig:replace_gan_with_mse}
\end{figure}

\textbf{Observation 1:} \textit{Hidden state in recurrent methods propagates the artifacts.}

In a unidirectional recurrent method, information in the hidden state serves two purposes. The first is to reconstruct the current frame with a reconstruction module, and the second is to propagate the previous information to the next frame. Therefore, once the output of the previous frame produces artifacts, the hidden state also contains such artifacts and further propagates to consequent frames. As shown in the $1^{st}$ and $3^{rd}$ rows in Fig.~\ref{fig:w_wo_hidden_state}, the contents of ellipse dotted box in $T_{t-1}$, $T_{t}$ retain and amplify the artifacts that appears in $T_{t-2}$. If the hidden state is not used (setting to $0$) and only the information of input image is used, the artifacts will be clearly alleviated. This phenomenon can be observed in the $2^{nd}$ and $4^{th}$ rows in Fig.~\ref{fig:w_wo_hidden_state}.
Directly discarding the hidden state, however, also lead to detail loss and unsharp edges ($2^{nd}$ and $4^{th}$ rows in Fig.~\ref{fig:w_wo_hidden_state}).

In a word, the hidden state not only enhances details but also propagates the artifacts, which is consistent with the observations in RealBasicVSR~\cite{chan2022investigating}.

\begin{figure}[t]
	\centering
	\includegraphics[width=1.0\columnwidth]{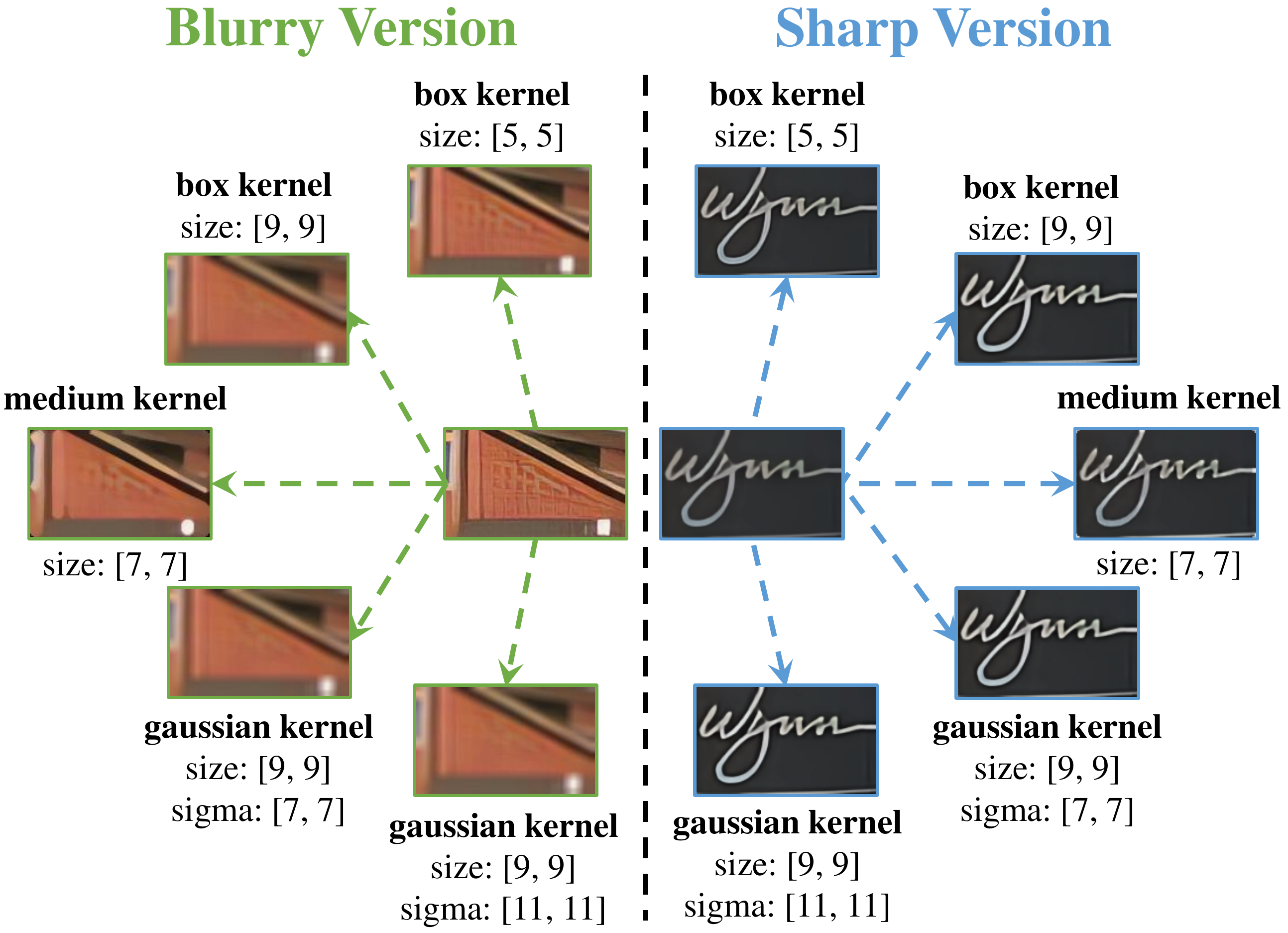}
	\caption{\textbf{Effects of different kernels.} For regions with artifacts, blur filters can mitigate artifacts. For regions with details, sharpening filters can further sharpen details.}
	\label{fig:filter}
\end{figure}

\noindent \textbf{Observation 2:} \textit{Replacing the hidden state in the GAN-based model with its corresponding hidden state in the MSE-based model can mitigate the artifacts.}

When the hidden state contains artifacts, although directly discarding the hidden state can alleviate the artifacts, it also leads to obvious details loss in other regions. Therefore, it is better to find a cleaner variant of the ``problematic'' hidden state which contains fewer artifacts and maintains its details. Typically, a GAN-based model is fine-tuned from an MSE-based model with a combination of MSE loss, perceptual loss~\cite{johnson2016perceptual} and GAN loss~\cite{goodfellow2014generative}. As shown in the $1^{st}$ row of Fig.~\ref{fig:replace_gan_with_mse}, the model trained with MSE loss tends to produce over-smooth results, thus the hidden state in the MSE-based model is cleaner and smooth.
Consequently, we wonder whether the artifacts of restored outputs can be mitigated if we replace the 'problematic' hidden state with the corresponding one in the MSE model. For better clarification, the hidden state in the MSE-based model and the hidden state in GAN-based model are referred to as MSE hidden state and GAN hidden state, respectively.

\begin{figure*}[t]
	\centering
	\includegraphics[width=1.9\columnwidth]{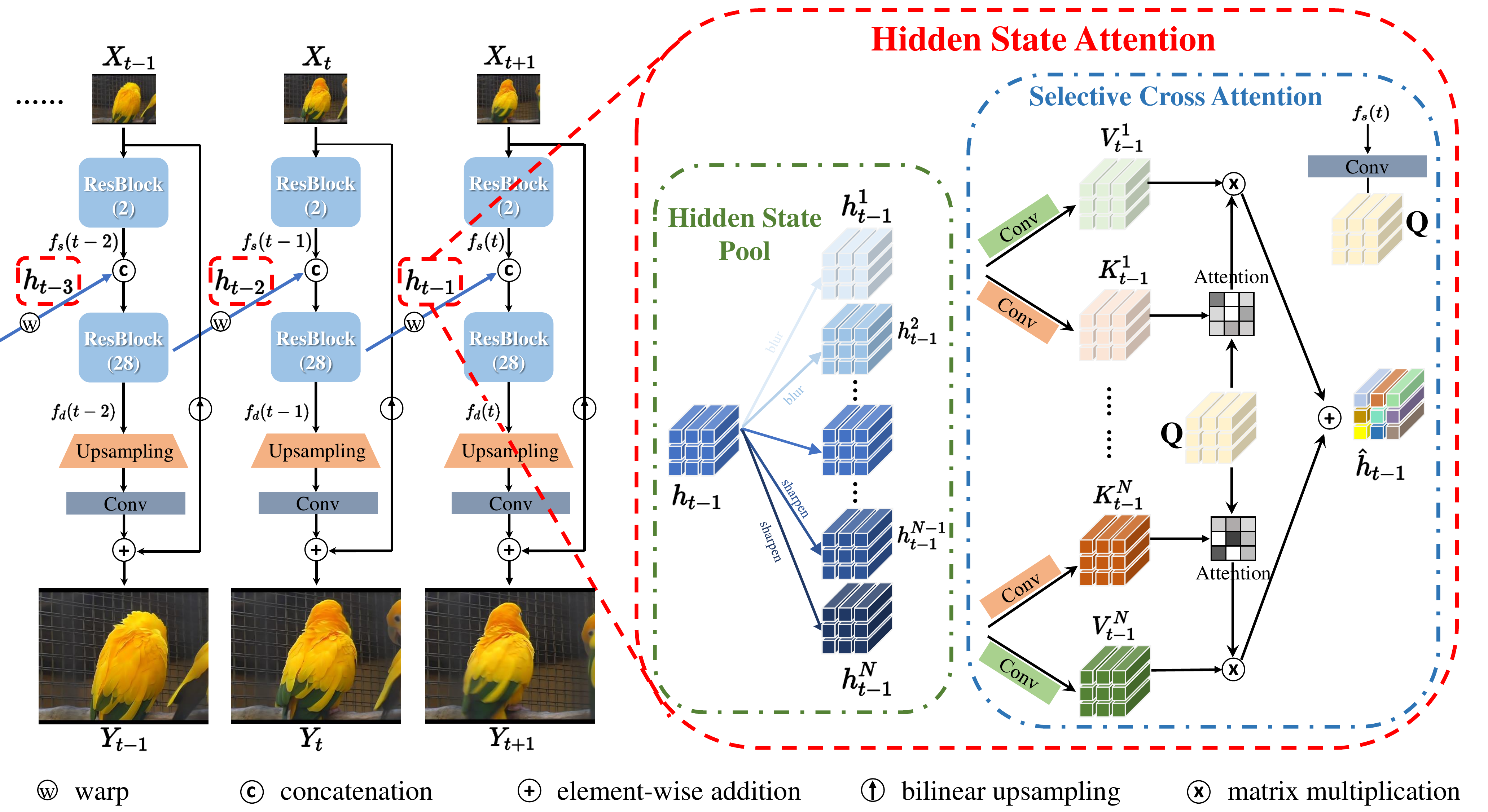}
	\caption{\textbf{Framework overview of FastRealVSR.} The left flowchart is a general unidirectional recurrent network. $f_{s}(t)$ and $f_{d}(t)$ represent the shallow and sharp feature of $\bm{X}_{t}$, respectively. $h_{t-1}$ is the hidden state that is propagated to restore $\bm{X}_{t}$. By adopting hidden state attention (HSA), $h_{t-1}$ is transformed into $\hat h_{t-1}$. HSA is composed of a hidden state pool and a selective cross attention module, as shown on the right side of the figure.}
	\label{fig:framework}
\end{figure*}

Fig.~\ref{fig:inference} illustrates the simplified diagram of a unidirectional recurrent network that uses both hidden state and input frame as inputs.
In Fig.~\ref{fig:inference}, we use a single color to represent a complete pre-trained inference model. To replace the GAN hidden state with MSE hidden state, firstly, we need to use $ \mathrm{Model_{mse}}$ (pre-trained MSE model) to infer the input sequence and store all generated hidden states. Then, we will infer the input sequence with $ \mathrm{Model_{combine}}$. $ \mathrm{Model_{combine}}$ is modified from $\mathrm{Model_{gan}}$ and the difference is that the GAN hidden state is replaced with the previously stored MSE hidden state. The qualitative comparison between these three models is shown in Fig.~\ref{fig:replace_gan_with_mse}. We can observe that \textbf{1)} from the comparison of the $2^{nd}$ and $3^{rd}$ rows (Fig.~\ref{fig:replace_gan_with_mse} (a)), the artifacts in  $ \mathrm{Model_{gan}} $ are reduced in $ \mathrm{Model_{combine}} $. \textbf{2)} directly replacing the GAN hidden state with the MSE hidden state could sometimes lose the necessary details (Fig.~\ref{fig:replace_gan_with_mse} (b)), compared against $\mathrm{Model_{gan}}$.

Although replacing the GAN hidden state with the MSE hidden state can mitigate artifacts, it is unpractical with two shortcomings. \textbf{1)} To obtain final results, it takes the double time since we need to run $\mathrm{Model_{mse}}$ and $\mathrm{Model_{combine}}$ in sequence. \textbf{2)} The output of $\mathrm{Model_{combine}}$ also loses many necessary details compared against $\mathrm{Model_{gan}}$. Therefore, it is preferable to directly manipulate the GAN hidden state, and construct a new hidden state that not only contains the smooth counterpart of the artifact region in the GAN hidden state, but also maintains the details of other regions.

\subsection{Framework}

Motivated by the above observations, we propose a hidden state attention (HSA) module to directly manipulate the hidden state and mitigate artifacts. An overview is illustrated in Fig.~\ref{fig:framework}.

\noindent \textbf{Formulation.} We first briefly introduce the pipeline of the general unidirectional recurrent VSR method. Let $X_{t}$ be the $t$-th image of the input sequence and $Y_{t}$ be the $t$-th frame of the restored sequence. Several residual blocks $\mathrm{RB_{1}}$ are used to extract the shallow feature of $X_{t}$:

\begin{equation}%
	f_{s}(t) = \mathrm{RB_{1}}(X_{t}).
\end{equation}

The hidden state from previous frame is first warpped with the estimated optical flow. The warpped hidden state is denoted as $h_{t-1}$. We use a pre-trained SPyNet~\cite{ranjan2017optical} to generate the optical flow. Then $h_{t-1}$ is concatenated with $f_{s}(t)$ and input to another set of residual blocks $\mathrm{RB_{2}}$ to get the deep feature $f_{d}(t)$:

\begin{equation}%
	f_{d}(t) = \mathrm{RB_{2}}(\mathrm{Concat}(h_{t-1}, f_{s}(t))),
\end{equation}

\noindent where $h_{t-1} \in \mathbb{R}^{H \times W \times C}$. $f_{d}(t)$ is then passed to an upsampling block $\mathrm{UP}$, and added with the bilinear upsampling of $X_{t}$ to generate $Y_{t}$:

\begin{equation}%
	Y_{t} = \mathrm{UP}(f_{d}(t)) + \mathrm{Bilinear}(X_{t}).
\end{equation}

For our method, we replace $h_{t-1}$ with $\hat h_{t-1}$ with
HSA:

\begin{equation}%
	\hat h_{t-1} = \mathrm{HSA}(h_{t-1}).
\end{equation}

\noindent where HSA is composed of a hidden state pool module and a selective cross attention module. The specific details of these two modules are described below.

\noindent \textbf{Hidden State Pool.} Hidden state pool is a container that gathers the blurry and sharp versions of $h_{t-1}$. The blurry version is obtained by a blurring operation, while the sharp version is obtained by the simplified sharpening algorithm (\ie unsharp masking (USM)). Specially, the sharp version is obtained by first subtracting the blurry version from the input, and then adding the residual to the input.

The above design is inspired by feature denoising~\cite{xie2019feature} that is adopted to clean feature maps for improving the adversarial robustness of convolutional networks. The effects of different blur operations are shown in Fig.~\ref{fig:filter}. Here, we use images to better illustrate the effects of blurry and sharpening filters. For the region with artifacts, we prefer its blurry version, while for the region with correct details, we prefer its sharp version.

In practical implementation, the operation of generating different versions of $h_{t-1}$ can be easily achieved by a convolution layer with a fixed kernel. Such computational operations are very cheap, and the computational cost is negligible. The process of generating the $i$-th blurry version of $h_{t-1}$ and the $j$-th sharp version of $h_{t-1}$ are formulated as:

\begin{equation}
	h_{t-1}^{i}= h_{t-1} \otimes \mathbf{k}^{i},
\end{equation}

\begin{equation}
	h_{t-1}^{j}= h_{t-1} + (h_{t-1} - h_{t-1} \otimes \mathbf{k}^{j}),
\end{equation}

\noindent where $\otimes$ denotes the convolving operation and $\mathbf{k}^{i}$, $\mathbf{k}^{j}$ are the blur kernels. The degree of blurring and sharpening can be controlled by adjusting the kernel size and variance (only specifically for the Gaussian filter). In our experiment, two mean filters and one Gaussian filter are adopted to generate blurry versions. We use two sharpening filters to generate the sharp versions. Detailed configurations of these filters are provided in the appendix.

\noindent \textbf{Selective Cross Attention.} After obtaining the hidden state pool $h_{t-1}^{i} \in \mathbb{R}^{H \times W \times C}$ ($1 \leq i \leq N$) ($N$ denotes the number of hidden state included in the hidden state pool), it is vital to aggregate a new hidden state that contains fewer artifacts. From the $2^{nd}$ and $4^{th}$ rows of Fig.~\ref{fig:w_wo_hidden_state}, we can observe that the artifacts in the output are largely mitigated if only the information of the input frame is utilized, indicating that the input feature contains fewer artifacts. Therefore, for the region with artifacts, the ``problematic'' hidden state has a large deviation from input features. Inspired by this, we propose a selective cross attention (SCA) module to aggregate a new hidden state $\hat h_{t-1}$ based on the attention between the hidden state pool $h_{t-1}^{i} \in \mathbb{R}^{H \times W \times C}$ ($1 \leq i \leq N$) and feature of input frame.

As shown in Fig.~\ref{fig:framework}, we first process $f_{s}(t)$ with a convolutional layer $H_{\mathrm{conv}}$, and get the query $Q \in \mathbb{R}^{H \times W \times C}$:

\begin{equation}
	Q=H_{\mathrm{conv}}(f_{s}(t)).
\end{equation}

For all $h_{t-1}^{i}$ ($1 \leq i \leq N$), we adopt two $3 \times 3$ convolutional layers to generate the key $K_{t-1}^{i}$ ($1 \leq i \leq N$) and value $V_{t-1}^{i}$ ($1 \leq i \leq N$), respectively. Then we calculate the similarity between each $K_{t-1}^{i}$ and $Q$ by the attention mechanism~\cite{vaswani2017attention,liu2021swin}. Based on the calculated attention maps, we aggregate $\hat h_{t-1}^{i}$ from $V_{t-1}^{i}$ ($1 \leq i \leq N$). The attention matrix is computed as

\begin{equation}
	\mathrm{SoftMax}([Q(K^{1})^{T};...;Q(K^{N})^{T}]) \otimes (V^{1};...;V^{N}).
\end{equation}

\section{Experiments}
\label{sec:experiments}

\subsection{Setup}

\noindent \textbf{Training Settings.} We train our FastRealVSR on the REDS~\cite{reds} dataset. The following degradation model is adopted to synthesize training data:
\begin{equation}%
	\bm{X} = [(\bm{Y} \circledast \bm{k}_{\sigma} + \bm{n}_{\delta})\downarrow_{r}]_{\mathrm{FFMPEG}}\ ,
\end{equation}

\noindent where $\bm{X}$ and $\bm{Y}$ are paired low-resolution and high-resolution sequences. The $\bm{k}$, $\bm{n}$, and $r$ are blur kernel, additive noise, and downsampling factor, respectively. We use a constant rate factor (\textit{crf}) to control the degree of FFMPEG compression. \textit{crf} uses a specific quality by adjusting the bitrates automatically. The sampling range of $\sigma$, $\delta$, and $\textit{crf}$ are $\{0.2 : 3\}, \{1 : 5\}$, and $\{18 : 35\}$, respectively.

\begin{figure}[t]
	\centering
	\includegraphics[width=1.0\columnwidth]{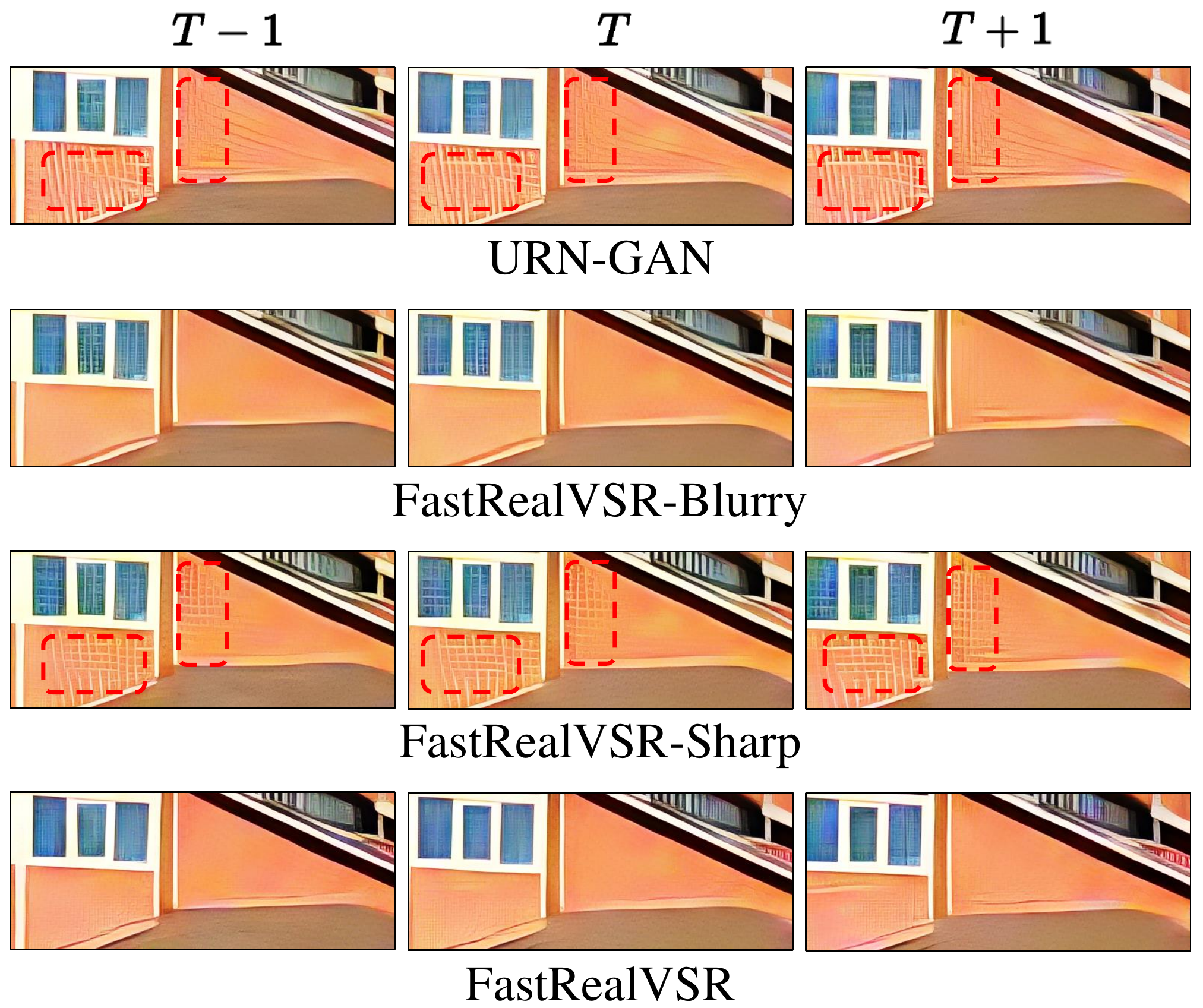}
	\caption{Results of URN-GAN, FastRealVSR-Blurry, FastRealVSR-Sharp and FastRealVSR. The outputs of URN-GAN and FastRealVSR-Sharp contain unpleasant artifacts. (\textbf{Zoom-in for best view})}
	\label{fig:training_hidden_state_cmp}
\end{figure}

Following RealBasicVSR, we load 15 frames as a sequence and flip them temporally in each iteration. The patch size of input LR frames is $64 \times 64$. The pre-trained SPyNet~\cite{ranjan2017optical} is used as the flow network and its weights are fixed in training. We employ Adam optimizer~\cite{kingma2014adam}. The training process is divided into two stages. 

\begin{figure*}[t]
	\centering
	\includegraphics[width=2.0\columnwidth]{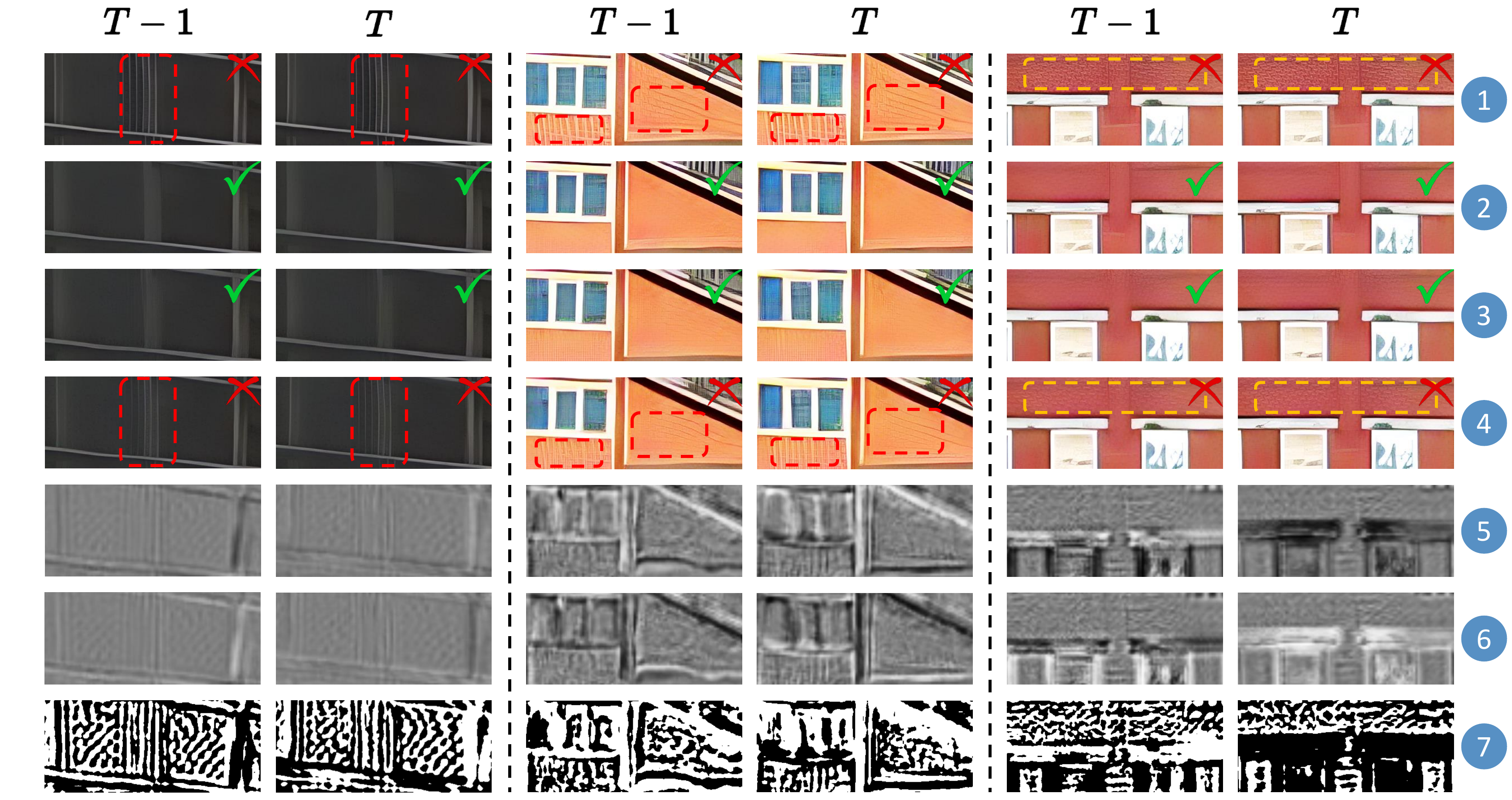}
	\vspace{-0.1cm}
	\caption{Analysis of hidden state pool and SCA. The outputs of URN-GAN contain obvious artifacts (\ding{172}). The artifacts produced in URN-GAN are largely alleviated in FastRealVSR (\ding{173}). When all hidden states in the hidden state pool are replaced with a single blurry hidden state, the artifacts are mitigated (\ding{174}). While the artifacts are retained when all hidden states are replaced with a single sharp hidden state (\ding{175}). The summed blurry/sharp attention maps are visualized in \ding{176} and \ding{177}, respectively. The binary map (\ding{178}) indicates the region where the value of \ding{176} is larger than \ding{177}. The regions marked with ellipse dotted boxes are artifacts. (\textbf{Zoom-in for best view})}
	\label{fig:filter_replacement_effect}
	\vspace{-0.2cm}
\end{figure*}

In the first stage, we adopt the Unidirectional Recurrent Network (URN) shown in Fig.~\ref{fig:framework}, and train it for $300$K iterations with the
$\textit{L}_{1}$ loss. The batch size and learning rate are set to $16$ and $10^{-4}$. In the second stage, we equip URN with the proposed HSA module to get the network FastRealVSR. We employ the pre-trained MSE model for initialization. Then we train FastRealVSR for $70$K iterations with a combination of $\textit{L}_{1}$ loss, perceptual loss~\cite{johnson2016perceptual} and GAN loss~\cite{goodfellow2014generative}, whose loss weights are set to ${1, 1,  5{\times}10^{-2}}$, respectively.
The batch size is set to $8$. The learning rate of the generator and discriminator is set to $5{\times}10^{-5}$ and $5{\times}10^{-4}$, respectively.
The exponential moving average (EMA)~\cite{yaz2018unusual} is adopted in training for more stable training and better performance. All experiments are implemented on 8 NVIDIA A100 GPUs with PyTorch~\cite{paszke2019pytorch}.

\noindent \textbf{Network Configuration.} In URN, the number of residual blocks that are used to extract the shallow and deep features is set to $2$ and $28$, respectively. The number of feature channels is $64$. In adversarial training, we adopt the discriminator of Real-ESRGAN~\cite{wang2021real}. For the hidden state pool, $3$ blur filters and $2$ sharpening filters are chosen to generate different hidden state variants. Detailed descriptions are provided in the appendix.

\noindent \textbf{Evaluation Dataset and Metrics.} We use VideoLQ~\cite{chan2022investigating} as test set and employ NIQE~\cite{niqe} and BRISQUE~\cite{brisque} to evaluate visual quality.

\subsection{Ablation Studies}

\noindent \textbf{Hidden State Pool.}
We use the GAN version of URN (URN-GAN) as the baseline model. Then we train FastRealVSR, FastRealVSR-Blurry and FastRealVSR-Sharp with the same settings. The only difference among them is the representation type of the hidden states in the hidden state pool. To be specific, FastRealVSR-Blur contains only blurry variants of hidden states, while FastRealVSR-Sharp contains only sharp ones. The quantitative comparison is listed in Tab.~\ref{tab:train_hidden_state}. The performance of FastRealVSR is significantly better than FastRealVSR-Blurry and FastRealVSR-Sharp, indicating that the information in the blurry and sharp representations of hidden states is complementary to each other.
Fig.~\ref{fig:training_hidden_state_cmp} shows the restored results of these four models. It can be observed that FastRealVSR and FastRealVSR-Blurry both produce outputs with fewer artifacts, whereas the outputs of URN-GAN and FastRealVSR-Blurry contain obvious artifacts. Besides, the edges of FastRealVSR are sharper than FastRealVSR-Blurry.

\noindent \textbf{Selective Cross Attention.} In SCA, each hidden state interacts with input features by attention. If the input feature is not used to guide the aggregation of hidden states and only a convolution layer is adopted to aggregate these hidden states, a performance drop is observed in Tab.~\ref{tab:sca}. It indicates that the guidance of input features is necessary when aggregating a new hidden state from a hidden state pool.

\begin{table}[th]
	\centering
	\caption{Ablation studies of different representation types in hidden state pool. FastRealVSR achieves the best performance.}
	\vspace{-0.2cm}
	\label{tab:train_hidden_state}\resizebox{0.5\linewidth}{!}{
		\begin{tabular}{c|c} \hline
			\textit{Network} & \textit{NIQE}$\downarrow$  \\ \hline
			URN-GAN                    & 3.8435 \\ \hline
			FastRealVSR-Blurry  & 3.8544 \\ \hline
			FastRealVSR-Sharp & 3.8210 \\ \hline
			FastRealVSR                &  3.7658 \\ \hline
		\end{tabular}
		\vspace{-0.5cm}
	}
\end{table}
\begin{table}[th]
	\centering
	\vspace{-0.7cm}
	\caption{Ablation study of SCA. With SCA, FastRealVSR achieves better performance.}
	\label{tab:sca}\resizebox{0.6\linewidth}{!}{
		\begin{tabular}{c|c|c} \hline
			\textit{Variant} & w/o SCA & FastRealVSR (w/ SCA)\\  \hline
			\textit{NIQE}$\downarrow$   &   3.8446      &  3.7658         \\ \hline
		\end{tabular}
	}
	\vspace{-0.5cm}
\end{table}

\begin{figure*}[t]
	\centering
	\vspace{-0.4cm}
	\includegraphics[width=2.0\columnwidth]{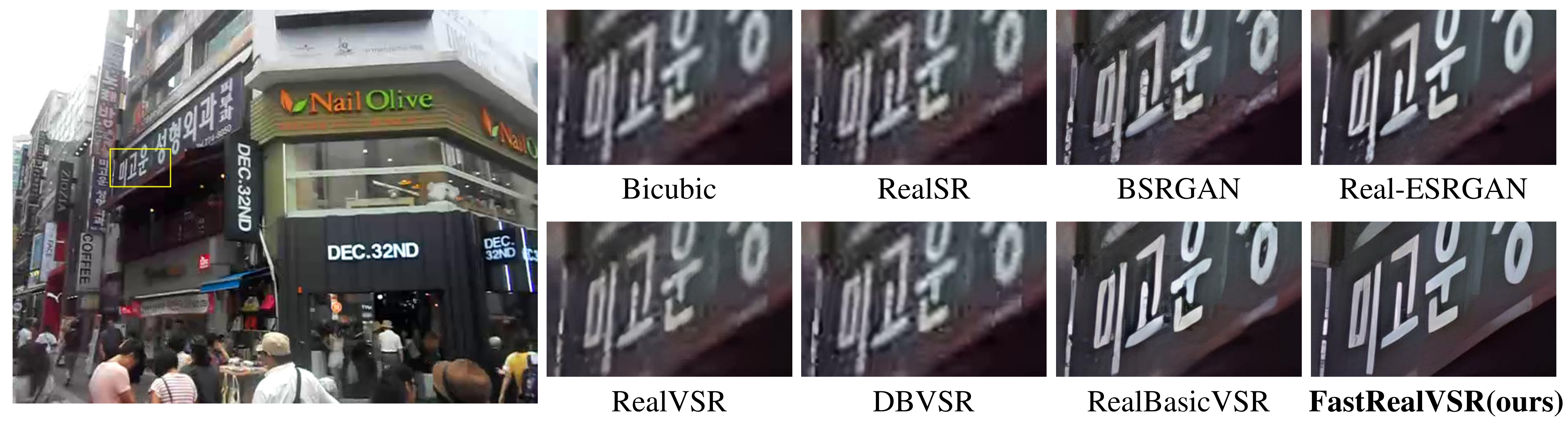}
	\caption{Qualitative comparison on the VideoLQ dataset for $\times 4$ video SR. (\textbf{Zoom-in for best view})}
	\vspace{-0.2cm}
	\label{fig:cmp}
\end{figure*}

\begin{table*}[t]
	\centering
	
	\caption{Quantitative comparison on VideoLQ dataset. FastRealVSR is capable of achieving a better speed-performance trade-off. Runtime is computed with an output size of $720 \times 1280$, with an NVIDIA V100 GPU. \redbf{Red} and \blueud{blue} indicates the best and second best performance. Following RealBasicVSR, metrics are computed on the Y-channel of the first, middle, and last frames of each sequence.}
	\vspace{-0.3cm}
	\label{tab:cmp}\resizebox{\linewidth}{!}{
		\begin{tabular}{l|c|c|c|c|c|c|c|c|cl} \hline
			& Bicubic & DAN    & RealSR & BSRGAN & Real-ESRGAN & RealVSR & DBVSR  & RealBasicVSR & \textbf{FastRealVSR} &  \\ \hline
			Params (M)   & -       & 4.3    & 16.7   & 16.7   & 16.7        & \redbf{2.7}     & 25.5   & 6.3          &  \blueud{3.8}         &  \\
			Runtime (ms) & -       & 185    & 149    & 149    & 149         & 1082    & 239    & \blueud{63}       & \redbf{30}          &  \\ \hline
			NIQE~\cite{niqe} $\boldsymbol{\downarrow}$       & 8.0049  & 7.1230 & 4.1482 & 4.2460 & 4.2091      & 8.0606  & 6.7866 & \blueud{3.7662}       & \redbf{3.7658}      &  \\
			BRISQUE~\cite{brisque} $\boldsymbol{\downarrow}$     & 54.899  & 51.563 & 30.542 & 30.213 & 32.103      & 54.988  & 50.936 & \redbf{29.030}       & \blueud{29.374}      &  \\ \hline
		\end{tabular}
	}
	\vspace{-0.2cm}
\end{table*}

\subsection{Analysis of Hidden State Attention}

The proposed HSA module aims to alleviate artifacts in the hidden state. Hence, we further validate the effectiveness of HSA for the ``problematic'' hidden state. When the previous output contains artifacts, the hidden state generated in URN-GAN contains artifacts and is a ``problematic'' hidden state (see Fig.~\ref{fig:training_hidden_state_cmp} and the $1^{st}$ row in Fig.~\ref{fig:filter_replacement_effect}). To get the ``problematic'' hidden state, we run URN-GAN and store all generated hidden states. We use $h_{t}$ ($1 \leq t \leq L - 1$) to refer to the ``problematic'' hidden state that produced by URN-GAN during the process of restoring low-resolution frame $\bm{X}_{t}$.
To test whether HSA in RealFastVSR has the capacity to mitigate artifacts, we replace the original hidden state in RealFastVSR with the ``problematic'' hidden state $h_{t-1}$ from URN-GAN when restoring $\bm{X}_{t}$.
From the $2^{nd}$ row in Fig.~\ref{fig:filter_replacement_effect}, we can observe that HSA has the ability to mitigate artifacts contained in the ``problematic'' hidden state.

\noindent \textbf{Investigation of Different Filters in Hidden State Pool.} We then investigate the respective effectiveness of blurry and sharp filters utilized in the hidden state pool. Specifically, we replace all hidden states in the hidden state pool with a single variant of hidden state, \ie, the hidden state pool contains identical hidden states after same filters. Note that the number of hidden states in the hidden state pool remains the same.
The results of adopting only a single blurry or a single sharp filter in hidden state pool are shown in the $3^{rd}$ and $4^{th}$ rows of Fig.~\ref{fig:filter_replacement_effect}, respectively. We can observe that blurry filters can mitigate artifacts but with the cost of detail loss, whereas sharp filters enhance both artifacts and details. Therefore, \textit{the combination of blurry filters and sharp filters can make a better trade-off between enhancing details and suppressing artifacts.}

\noindent \textbf{Visualization of Attention Map in SCA.} We also visualize the attention map in the selective cross attention (SCA) module to understand how the information in different representations of hidden state is selected based on the input feature. For better visualization, blurry and sharp attention maps are summed together, respectively, resulting in two maps. The visualization of these two maps is shown in the $5^{th}$ and $6^{th}$ rows in Fig.~\ref{fig:filter_replacement_effect}.
We can observe that for different versions (\ie, blurry or sharp) of $h_{t-1}$, SCA indeed learns to select the proper information with end-to-end training.
Besides, we compare these two maps and get the binary map shown in the $7^{th}$ row in Fig.~\ref{fig:filter_replacement_effect}. The binary map indicates the region where the value of blurry attention map is larger than the sharp one. It can be observed that \textit{for the regions with artifacts, SCA prefers the blurry representations.}

\subsection{Comparison with State-of-the-Art Methods}
We compare our FastRealVSR method with seven algorithms: DAN~\cite{dan}, RealSR~\cite{real-sr}, BSRGAN~\cite{bsrgan}, Real-ESRGAN~\cite{wang2021real}, RealVSR~\cite{realvsr}, DBVSR~\cite{dbvsr} and RealBasicVSR~\cite{chan2022investigating} on the VideoLQ~\cite{chan2022investigating} dataset. VideoLQ consists of $50$ real-world low-quality videos that are collected from various video hosting sites such as Flickr and YouTube.

The quantitative results on VideoLQ are shown in Tab.~\ref{tab:cmp}. Compared with other methods, FastRealVSR achieves a better speed-performance trade-off. In particular, compared to RealBasicVSR, FastRealVSR achieves comparable performance with $2$ $\times$ faster speed and smaller model size. Qualitative results are presented in Fig.~\ref{fig:cmp}. Compared to previous methods, FastRealVSR can effectively remove annoying artifacts and preserve sharp details.

\section{Acknowledgments}
This work was supported in part by the National Natural Science Foundation of China under Grant (62276251,U1913210), the Joint Lab of CAS-HK,  in part by  the Shanghai Committee of Science and Technology, China (Grant No. 20DZ1100800).

\newpage
\appendix
\section{Appendix}

In this section, we first provide detailed descriptions of adopted blur filters and sharpening filters. Then we show more quantitative and qualitative comparisons with previous works.

\begin{figure*}[t]
	\centering
	\includegraphics[width=2.0\columnwidth]{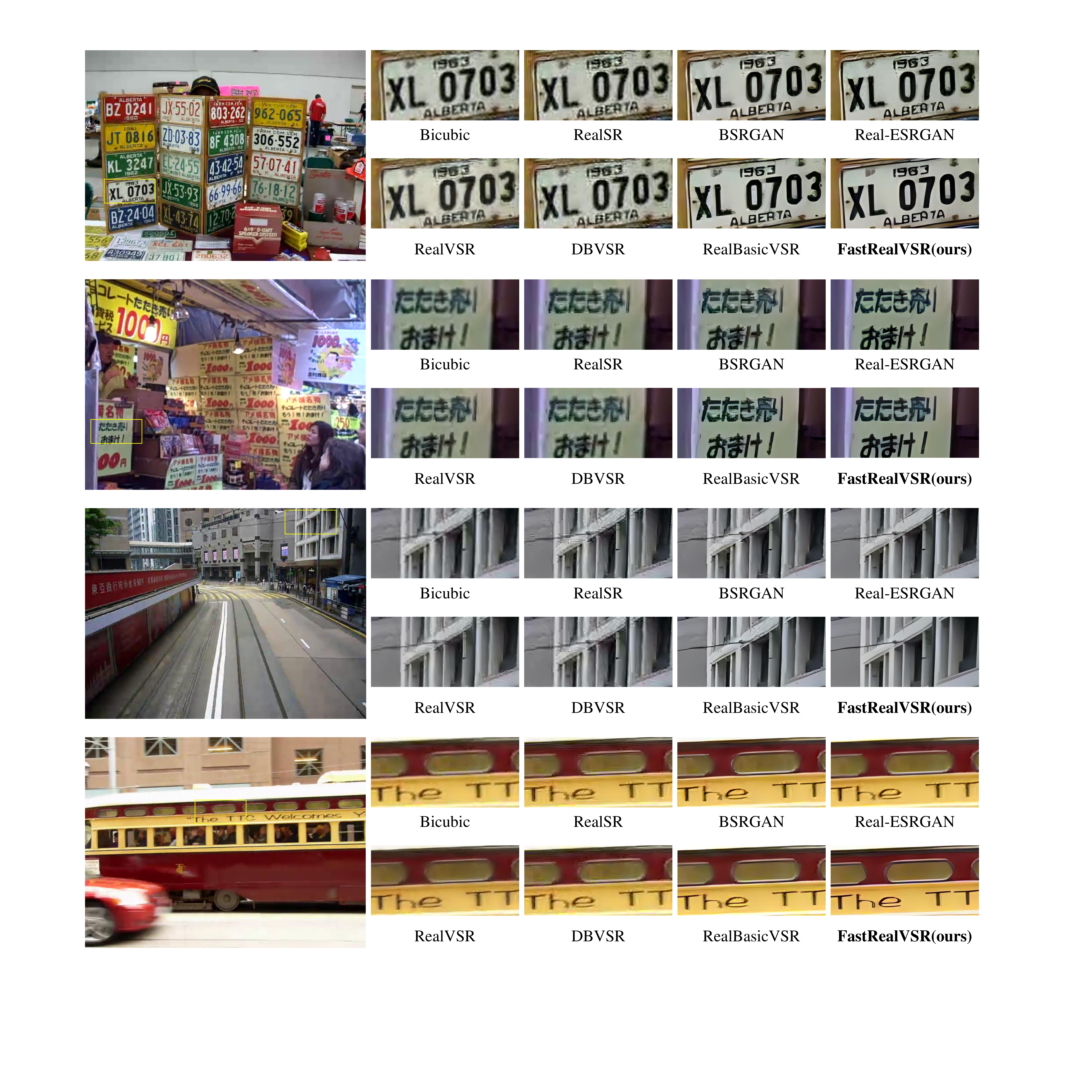}
	\vspace{-0.4cm}
	\caption{Qualitative comparison on the VideoLQ dataset for $\times 4$ video SR. Compared against other methods, FastRealVSR can better remove the noise around texts and buildings, with clearer edges. (\textbf{Zoom-in for best view})}
	\label{fig:cmp1}
\end{figure*}

\subsection{Details of Adopted Filters}

\textbf{Blurry filters:} In our experiment, we adopt two mean filters and one Gaussian filter to generate three blurry hidden states. Their specific kernels are listed in Tab.~\ref{tab:blurry_filters}.

\begin{table}[th]
	\centering
	\caption{The specific kernel of adopted blur filters.}
	\label{tab:blurry_filters}\resizebox{0.7\linewidth}{!}{
		\begin{tabular}{|c|c|c|} \hline
			Index & Filter type & Filter weight \\ \hline
			\ding{172} & Mean filter & $\begin{bmatrix}
				\frac{1}{9} & \frac{1}{9} & \frac{1}{9} \\
				\frac{1}{9} & \frac{1}{9} & \frac{1}{9} \\
				\frac{1}{9} & \frac{1}{9} & \frac{1}{9}
			\end{bmatrix}$ \\ \hline
			\ding{173} & Mean filter & $\begin{bmatrix}
				\frac{1}{25} & \frac{1}{25} & \frac{1}{25} & \frac{1}{25} & \frac{1}{25}\\
				\frac{1}{25} & \frac{1}{25} & \frac{1}{25} & \frac{1}{25} & \frac{1}{25}\\
				\frac{1}{25} & \frac{1}{25} & \frac{1}{25} & \frac{1}{25} & \frac{1}{25} \\
				\frac{1}{25} & \frac{1}{25} & \frac{1}{25} & \frac{1}{25} & \frac{1}{25} \\
				\frac{1}{25} & \frac{1}{25} & \frac{1}{25} & \frac{1}{25} & \frac{1}{25} 
			\end{bmatrix}$ \\ \hline
			\ding{174} & Gaussian filter & $\begin{bmatrix}
				0.1108 & 0.1113 & 0.1108 \\
				0.1113 & 0.1117 & 0.1113 \\
				0.1108 & 0.1113 & 0.1108
			\end{bmatrix}$ \\ \hline
		\end{tabular}
	}
\end{table}

\noindent \textbf{Sharp filters:} For generating the sharp versions of the hidden state, we adopt two sharp filters. Their specific kernels are listed in Tab.~\ref{tab:sharp_filters}.

\begin{table}[th]
	\centering
	\caption{The specific kernel of adopted sharp filters.}
	\label{tab:sharp_filters}\resizebox{0.9\linewidth}{!}{
		\begin{tabular}{|c|c|c|} \hline
			Index & Filter type & Filter weight \\ \hline
			\ding{172} & Gaussian filter & $\begin{bmatrix}
				0.1096 & 0.1118 & 0.1096 \\
				0.1118 & 0.1141 & 0.1118 \\
				0.1096 & 0.1118 & 0.1096 
			\end{bmatrix}$ \\ \hline
			\ding{173} & Gaussian filter & $\begin{bmatrix}
				0.0369 & 0.0392 & 0.0400 & 0.0392 & 0.0369 \\
				0.0392 & 0.0416 & 0.0424 & 0.0416 & 0.0392 \\
				0.0400 & 0.0424 & 0.0433 & 0.0424 & 0.0400 \\
				0.0392 & 0.0416 & 0.0424 & 0.0416 & 0.0392 \\
				0.0400 & 0.0424 & 0.0433 & 0.0424 & 0.0400 
			\end{bmatrix}$ \\ \hline
			
		\end{tabular}
	}
\end{table}

\subsection{Comparison with other methods}

We compare our FastRealVSR method with seven algorithms: DAN~\cite{dan}, RealSR~\cite{real-sr}, BSRGAN~\cite{bsrgan}, Real-ESRGAN~\cite{wang2021real}, RealVSR~\cite{realvsr}, DBVSR~\cite{dbvsr} and RealBasicVSR~\cite{chan2022investigating} on the VideoLQ~\cite{chan2022investigating} dataset. VideoLQ consists of $50$ real-world low-quality videos that are collected from various video hosting sites such as Flickr and YouTube.

To better evaluate the performance of these methods, we adopt another metric MANIQA~\cite{yang2022maniqa} to evaluate the visual quality. MANIQA is the champion of the NTIRE 2022 Perceptual Image Quality Assessment Challenge and its performance outperforms state-of-the-art methods by a large margin. Therefore, this metric is more convincing for comparison of the results of these methods. As mentioned in RealBasicVSR~\cite{chan2022investigating}, to save the time, NIQE and BRISQUE metrics are adopted to evaluate the quality score of the first, middle and last frames of each sequence. This evaluation strategy can not comprehensively evaluate an algorithm. For MANIQA, we adopt two evaluation strategies, the first is to evaluate three frames as RealBasicVSR, while the second is to evaluate all frames of each sequence. Both two evaluation strategies evaluate the quality of frames on the RGB channel.

The quantitative results on VideoLQ are shown in Tab.~\ref{tab:cmp}. For RealBasicVSR, we use the official model and testing strategy to test VideoLQ dataset, the evaluation result of PI is different from the value that represented in their paper. From this table, we can observe that FastRealVSR achieves comparable performance with RealBasicVSR, with $2$ $\times$ faster speed and smaller model size. For the state-of-the-art image quality assessment, MANIQA, we can observe that the performance of FastRealVSR outperforms other methods.

Qualitative results are presented in Fig.~\ref{fig:cmp1}. Compared to previous methods, FastRealVSR can effectively remove annoying artifacts and preserve sharp details. 

\begin{table*}[t]
	\centering
	
	\caption{Quantitative comparison on VideoLQ dataset. FastRealVSR is capable of achieving a better speed-performance trade-off. Runtime is computed with an output size of $720 \times 1280$, with an NVIDIA V100 GPU. \redbf{Red} and \blueud{blue} indicates the best and second best performance. Following RealBasicVSR, NIQE and BRISQUE are computed on the Y-channel of the first, middle, and last frames of each sequence. For MANIQA, we adopt two evaluation strategies, the first is to evaluate three frames as what RealBasicVSR does, while the second is to evaluate all frames of each sequence. Both two evaluation strategies evaluate the quality of frames on the RGB channel.}
	\vspace{-0.3cm}
	\label{tab:cmp}\resizebox{\linewidth}{!}{
		\begin{tabular}{l|c|c|c|c|c|c|c|c} \hline
			& Bicubic & DAN    & RealSR & BSRGAN & Real-ESRGAN & RealVSR  & RealBasicVSR & \textbf{FastRealVSR}  \\ \hline
			Params (M)   & -       & 4.3    & 16.7   & 16.7   & 16.7        & \redbf{2.7}     & 6.3       &  \blueud{3.8}          \\
			Runtime (ms) & -       & 185    & 149    & 149    & 149         & 1082    & \blueud{63}       & \redbf{30}           \\ \hline
			NIQE~\cite{niqe} $\boldsymbol{\downarrow}$ (\textit{3 frames})       & 8.0049  & 7.1230 & 4.1482 & 4.2460 & 4.2091      & 8.0606  & \blueud{3.7662}    & \redbf{3.7658}     \\
			PI~\cite{pi} $\boldsymbol{\downarrow}$  (\textit{3 frames})    & 7.6017  & 6.8942 & 4.2648 & 4.2644 & 4.2492      & 7.7824  & \blueud{3.9152*}   &   \redbf{3.8975}      \\
			BRISQUE~\cite{brisque} $\boldsymbol{\downarrow}$ (\textit{3 frames})    & 54.899  & 51.563 & 30.542 & 30.213 & 32.103      & 54.988  & \redbf{29.030} & \blueud{29.374}     \\ 
			MANIQA~\cite{yang2022maniqa}  $\boldsymbol{\uparrow}$ (\textit{\textbf{3} frames}) & 0.3019 & 0.2529 & 0.2896  & 0.4265 & 0.4347 & 0.2560 & \blueud{0.4387} & \redbf{0.4483} \\ \hline
			MANIQA~\cite{yang2022maniqa}  $\boldsymbol{\uparrow}$ (\textit{\textbf{all} frames}) & 0.3011 & 0.2522  & 0.2827 & 0.4233 & 0.4302 & 0.2567 & \blueud{0.4385} & \redbf{0.4637} \\ \hline
		\end{tabular}
	}
\end{table*}

\clearpage
\clearpage
\newpage
\bibliography{aaai23}

\end{document}